\documentclass[a4paper,11pt]{article}

\usepackage{fourier}
\usepackage{color}
 \usepackage{graphicx}
\usepackage{url}
\usepackage[affil-it]{authblk}
\usepackage{amsmath}
\usepackage{wrapfig}
\usepackage{xspace}

\usepackage[T1]{fontenc}
\usepackage{times}

\usepackage{epsfig}
\usepackage{graphicx}
\usepackage{amsmath}
\usepackage{amssymb}
\usepackage{color}
\usepackage{tabularx, booktabs}
\usepackage{hyperref} 

\usepackage{subcaption}
\usepackage{multirow}

\usepackage{float}

\usepackage{footmisc} 

\begin{document}

\title{A Study of Efficient Light Field\\ Subsampling and Reconstruction Strategies}


\author{Yang Chen}
\author{Martin Alain}
\author{Aljosa Smolic}

\affil{
 V-SENSE project\\
 Graphics Vision and Visualisation group (GV2)\\
 Trinity College Dublin
 \thanks{This publication has emanated from research conducted with the financial support of Science Foundation Ireland (SFI) under the Grant Number 15/RP/2776. We also gratefully acknowledge the support of NVIDIA Corporation with the donation of the Titan Xp GPU used for this research.}
}

\date{}
\maketitle
\thispagestyle{empty}

\vspace{-0.1cm}

\begin{abstract}
Limited angular resolution is one of the main obstacles for practical applications of light fields. Although numerous approaches have been proposed to enhance angular resolution, view selection strategies have not been well explored in this area. In this paper, we study subsampling and reconstruction strategies for light fields.
First, different subsampling strategies are studied with a fixed sampling ratio,
such as row-wise sampling, column-wise sampling, or their combinations. Second, several strategies are explored to reconstruct intermediate views from four regularly sampled input views. The influence of the angular density of the input is also evaluated.
We evaluate these strategies on both real-world and synthetic datasets, and optimal selection strategies are devised from our results. These can be applied in future light field research such as compression, angular super-resolution, and design of camera systems.
\end{abstract}

\section{Introduction}
\label{sec:intro}
\begin{wrapfigure}{r}{0.5\textwidth}
  \vspace{-20pt}
  \begin{center}
    \includegraphics[width=0.48\textwidth, height=0.28\textwidth]{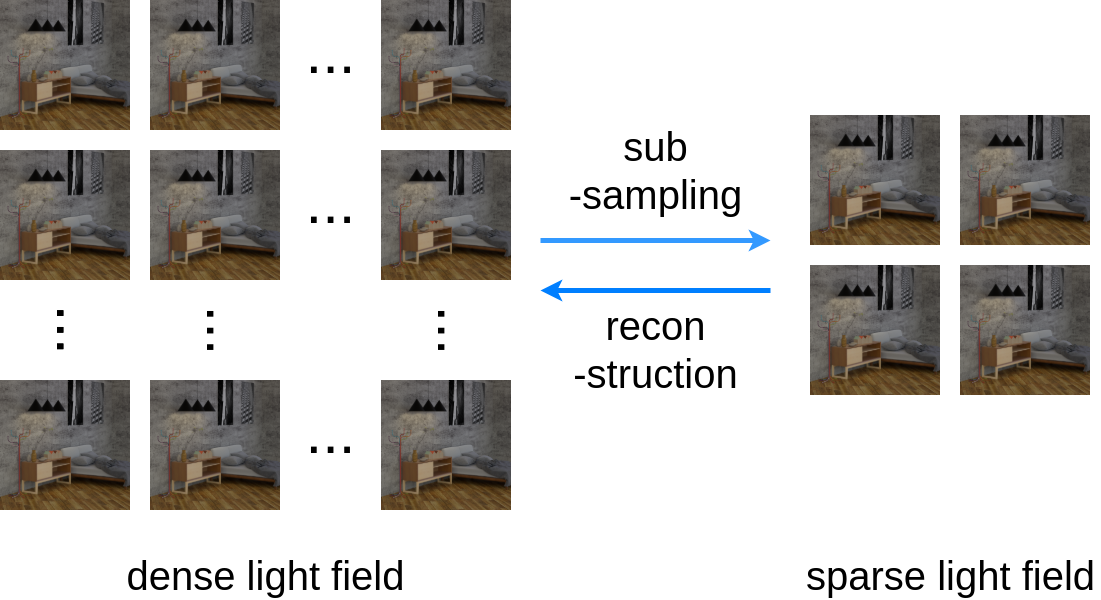}
\end{center}
\vspace{-20pt}
  \caption{\small{Subsampling and reconstruction of a two-plane representation light field.}}
  \label{fig:intro}
  \vspace{-0pt}
\end{wrapfigure}

A light field (LF) is described as all light rays passing through a given 3D volume in the pioneer work of Levoy et al.~\cite{Levoy1996}. Different from traditional 2D imaging systems, a 4D LF imaging system captures not only two spatial but also two additional angular dimensions. The two plane representation of light fields is adopted in this paper, which is represented as a collection of views taken from several view points parallel to a common plane, as shown in Figure~\ref{fig:intro}. Another common representation of light fields are Epipolar Plane Images (EPI), which are 2D slices of the 4D light field obtained by fixing one spatial and one angular dimension. 

Dense light fields are advantageous for various applications including virtual reality, augmented reality and image-based rendering~\cite{yu2017light, wu2017overview}. However, dense light fields are usually difficult to capture in practice and a trade-off has to be found between spatial and angular resolution. For example, camera arrays usually have good spatial resolution but sparse angular sampling, while the opposite is usually the case for plenoptic cameras. 

Spatial super-resolution of light fields is a widely studied problem and many methods have been proposed with impressive results~\cite{alain2018light, rossi2017graph, yoon2015learning}. 
As for the angular resolution enhancement in light fields, one common solution is to apply view synthesis methods on LFs~\cite{yoon2015learning, kalantari2016learning, wang2018end, vagharshakyan2017light}. However, only limited attention has been paid to view selection strategies and previous work usually arbitrarily selects one fixed input pattern, such as along one angular row or within a $N \times N$ square matrix of views.

In this paper, we investigate light field subsampling and reconstruction strategies and evaluate their performance. We identify three issues to face:
First, a benchmark method is required to compare these strategies. State-of-the-art (SOTA) light field subsampling and reconstruction methods are reviewed and evaluated experimentally, and the best performing approach is chosen as the benchmark for following experiments.
Second, since LFs contain huge amounts of data, which poses challenges for storage and transmission, efficient subsampling methods are needed for LFs. Given a dense LF as input, we investigate which subsampling strategy can produce the best reconstruction results while keeping a satisfying sampling ratio.
Third, since dense LFs are expensive to capture, reconstructing a dense LF from sparse input becomes an important and challenging topic. Here we investigate different strategies and different levels of sparsity for the task of reconstructing a LF from corner views. Eventually, we derive optimal strategies for each problem based on our results. 

This paper is organized as follows. In Section~\ref{sec:related}, we review existing methods for light field angular super-resolution and more general techniques for view synthesis and video frame interpolation. In Section~\ref{sec:study}, a benchmark method is selected and the experiments for evaluation of subsampling and reconstruction strategies are described in detail. Then, these designs are evaluated and compared in Section~\ref{sec:results} after applying the selected benchmark method. Finally, we present our conclusions in Section~\ref{sec:conclusion}.

\begin{figure}[t]
    \hspace{1.3cm}
    \begin{subfigure}{0.25\textwidth}
        \centering
        \includegraphics[width=\linewidth]{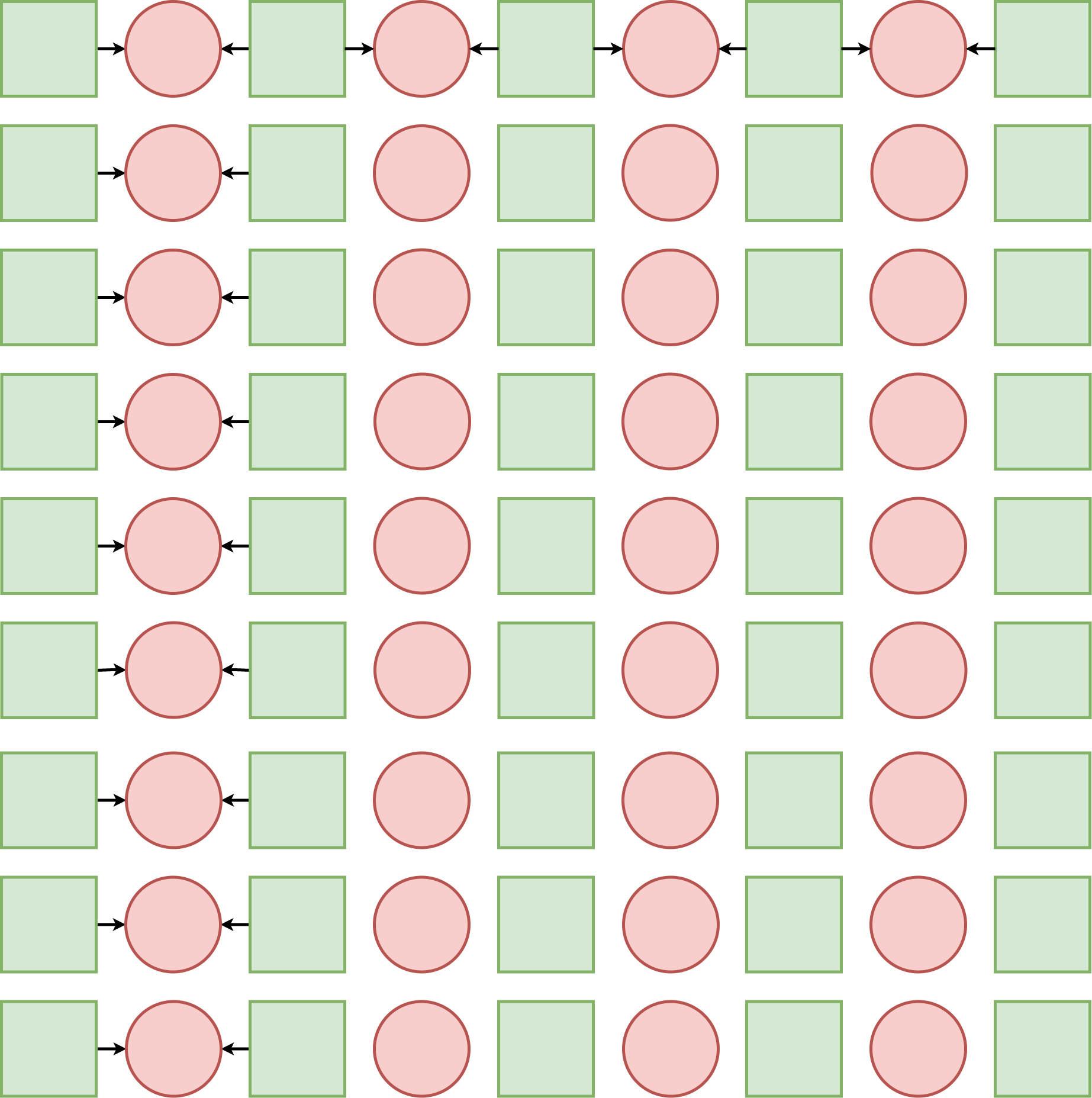} 
        \caption{\small{row-wise}}
        \label{fig:fullLF_subimg1}
    \end{subfigure}
    \hspace{0.5cm}
    \begin{subfigure}{0.25\textwidth}
        \centering
        \includegraphics[width=\linewidth]{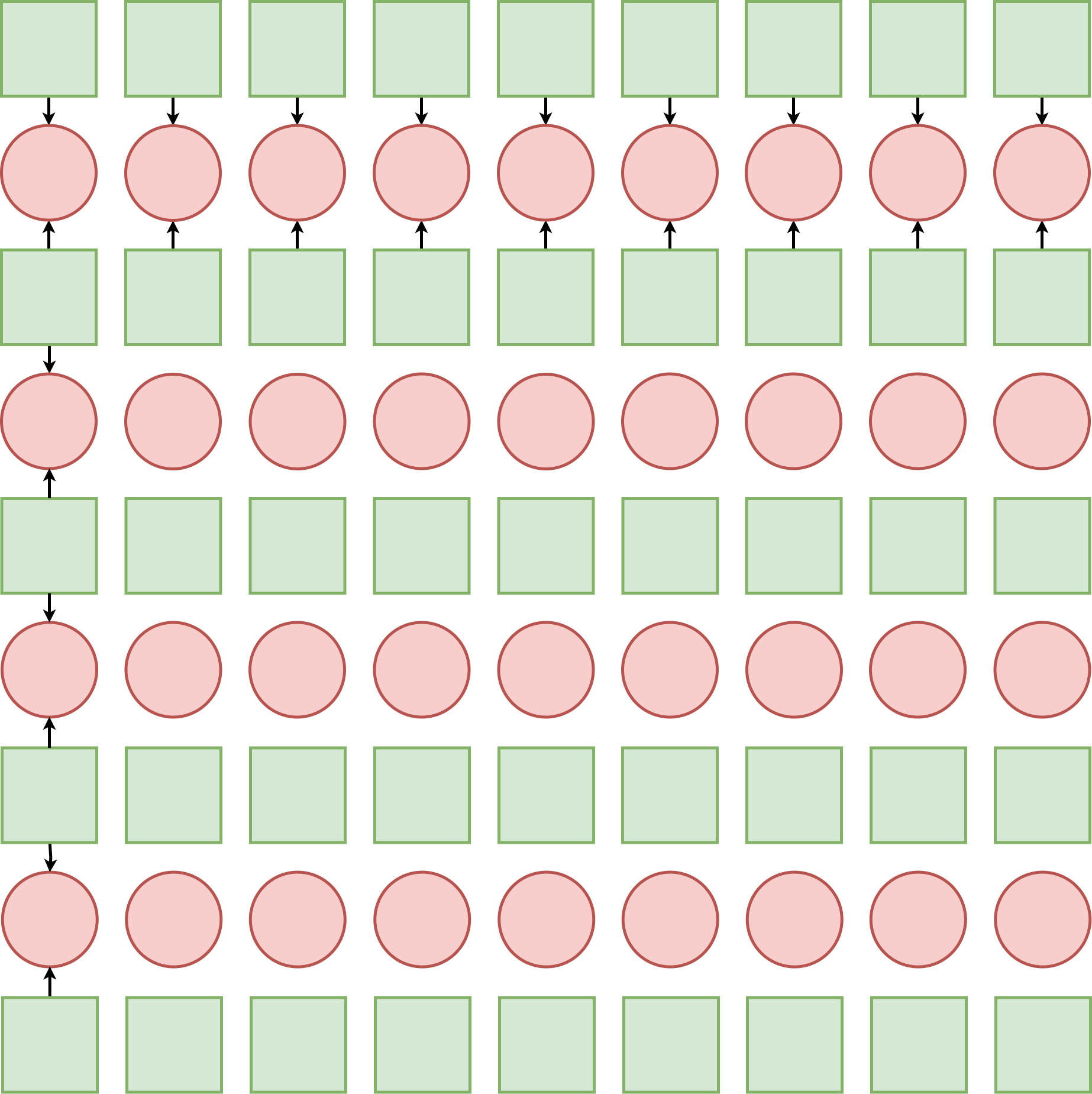}
        \caption{\small{column-wise}}
        \label{fig:fullLF_subimg2}
    \end{subfigure}
    \hspace{0.5cm}
    \begin{subfigure}{0.25\textwidth}
        \centering
        \includegraphics[width=\linewidth]{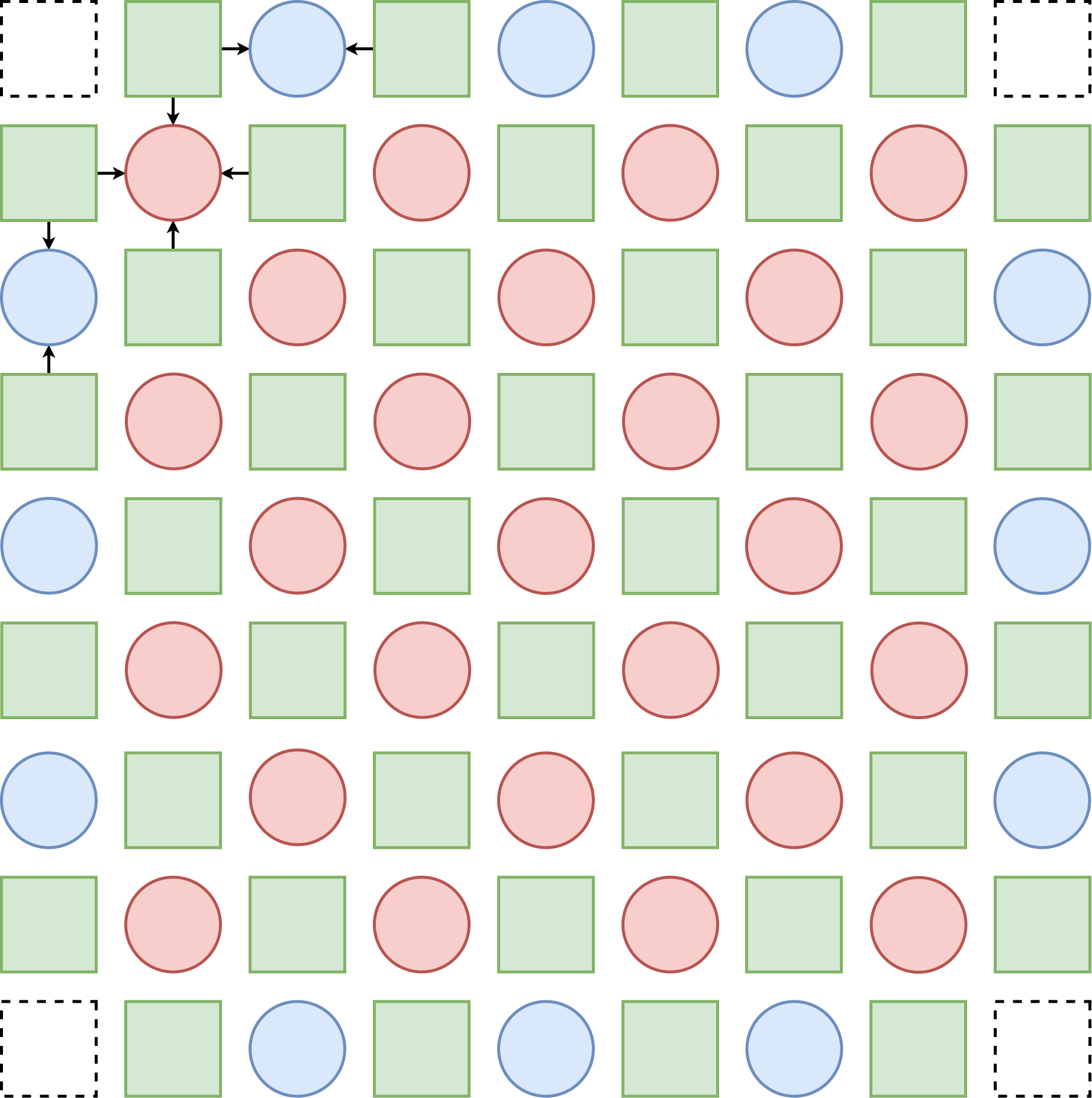}
        \caption{\small{checkerboard}}
        \label{fig:fullLF_subimg3}
    \end{subfigure}
     
    \caption{\small{Three basic subsampling strategies. The squares are sampled views, the circles are reconstructed views and the dashed squares are unused views. Blue circles in (c) are reconstructed by row-wise or column-wise from two adjacent views to complete the LF.}}
    \label{fig:fullLF}
    \vspace{-0.6cm}
\end{figure}

\section{Related Work}
\label{sec:related}
\textbf{Angular Super-resolution for Light Fields}
Wanner and Goldluecke proposed a variational framework to generate novel views from sparse input views, of which ghost artifacts can still be observed on the synthesized results~\cite{wanner2013variational}. Densely sampled light fields have small disparity between adjacent views, which makes them very suitable for frequency domain analysis. Therefore, Shi et al. proposed to reconstruct the dense light field by optimizing frequency coefficients in the continuous 2D Fourier domain~\cite{shi2014light}. A more advanced shearlet transform is adapted in~\cite{vagharshakyan2017light} on Epipolar Plane Images. However, wide EPI input is required by these Fourier methods, which is not feasible for some applications.

The strength of deep learning based approaches has been demonstrated in sevaral LF super-resolution methods. A first example designed a multi-stream network for light field spatial and angular super-resolution \cite{yoon2015learning}. Kalantari et al. proposed a learning based framework for light field view synthesis, which requires depth estimation as an intermediate step~\cite{kalantari2016learning}.
A sparse input consisting of the corner views of a dense light field is fed into a convolutional neural network to synthesize the original intermediate views.
Wu et al. re-modeled light field angular super-resolution as a detail restoration problem in 2D EPI space (LFEPI)~\cite{wu2017epi}. A blur-restoration-deblur framework is built to process EPIs of a sparsely sampled LF and to recover the angular detail with a convolutional neural network. To utilize inherent consistency of the LF, Wang et al. introduced a pseudo 4D convolution by combining a 2D convolution on EPIs and a sequential 3D convolution~\cite{wang2018end}. While these methods presented innovative ideas and good results, fixed subsampling and reconstruction strategies are applied without investigating and optimising alternatives.

\textbf{View Synthesis and Video Frame Interpolation}
Existing depth image-based rendering and video frame interpolation methods can be directly applied to the LF angular super-resolution problem. A deep architecture was introduced by Flynn et al. to synthetize novel views from wide disparity real-world inputs~\cite{flynn2016deepstereo}. Niklaus et al. proposed pairs of spatially individual 1D kernels, which are estimated from a trained convolutional neural network, to estimate motion and color interpolation within one stage (SepConv)~\cite{niklaus2017video}. 

\begin{table}[!t]
\footnotesize
    \begin{center}
    \caption{Comparison between SOTA LF interpolation methods}
    \label{tab:SOTA}
    \vspace{-0.2cm}
    \resizebox{0.7\textwidth}{!}{
    \begin{tabular}{c|c|c|c|c}
        \multirow{2}{*}{row-wise} &  \multicolumn{3}{c|}{PSNR/SSIM} \\
        \cline{2-4}
        & \text{Mean} & \text{HCI} & \text{Stanford} \\
        \hline
        \text{Linear} & 32.37/0.9464 & 31.82/0.9501 & 33.47/0.9390 \\
        \hline
        \text{Shearlet~\cite{vagharshakyan2017light}} & 34.92/0.9592 & 35.78/0.9750 & 33.21/0.9277 \\
        \hline
        \text{LFEPI~\cite{wu2017epi}} & 37.39/0.9451 & 37.51/0.9420 & 37.15/0.9515 \\
        \hline
        \text{SepConv~\cite{niklaus2017video}} & \color{red}{\textbf{38.94}}/\color{red}{\textbf{0.9910}} & \color{red}{\textbf{39.38}}/\color{red}{\textbf{0.9945}} & \color{red}{\textbf{38.07}}/\color{red}{\textbf{0.9839}}
    \end{tabular}
    }
    \end{center}
    \vspace{-0.4cm}
\end{table}

\begin{table}[!t]
\footnotesize
    \begin{center}
    \caption{Comparison of pretrained, fine-tuned and retrained SepConv for row-wise interpolation}
    \label{tab:SepConv}
    \begin{tabular}{c|c|c|c}
        row-wise & pretrained & Fine-tuned & Retrained \\
        \hline
        \text{PSNR(dB)} & 38.08 & \color{red}{\textbf{39.74}} & 39.41 \\
        \hline
        \text{SSIM} & 0.9893 & \color{red}{\textbf{0.9925}} & 0.9923
    \end{tabular}
    \end{center}
    \vspace{-0.4cm}
\end{table}




\section{Study of efficient subsampling and reconstruction strategies}
\label{sec:study}
\begin{figure}[t]
    \hspace{2.5cm}
    \begin{minipage}{.2\linewidth}
        \centering
        \subcaption{\small{2D H-V}}
        \includegraphics[width=\linewidth, scale=.15]{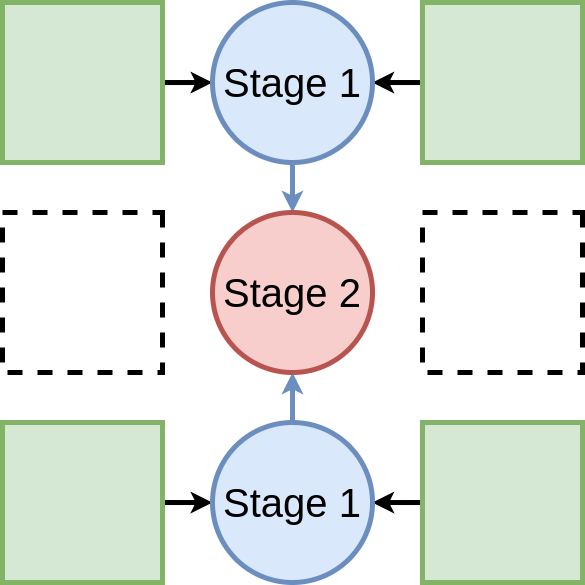}
    \end{minipage}
    \hspace{0.5cm}
    \begin{minipage}{.2\linewidth}
        \centering
        \subcaption{\small{2D V-H}}
        \includegraphics[width=\linewidth, scale=.15]{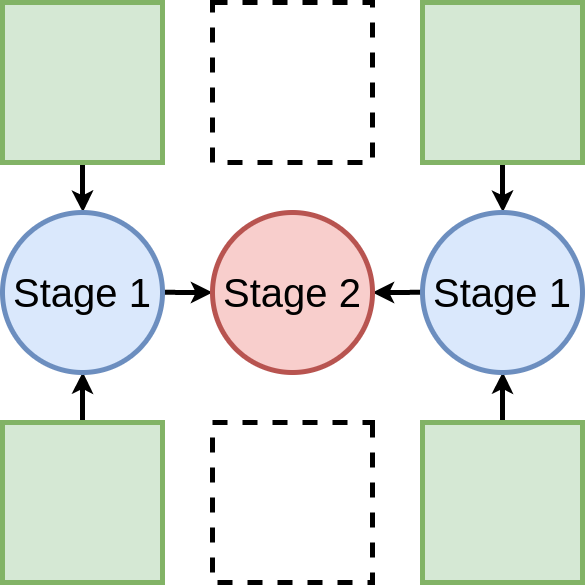}
    \end{minipage}
    \hspace{0.5cm}
    \begin{minipage}{.2\linewidth}
        \centering
        \subcaption{\small{4D omni}}
        \includegraphics[width=\linewidth, scale=.15]{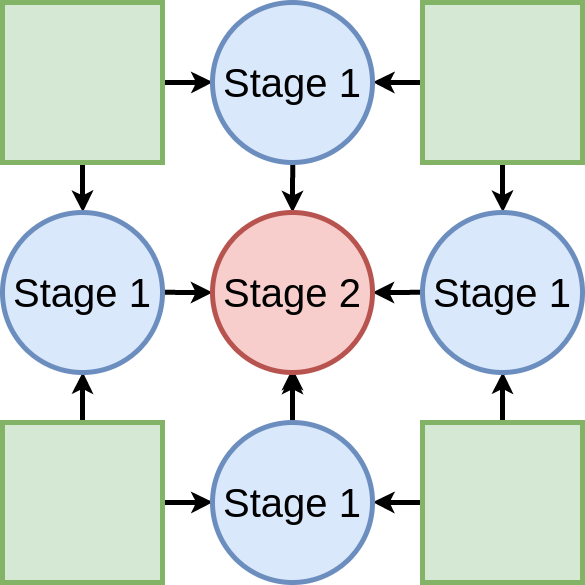}
    \end{minipage}\par\medskip
    \hspace{2.5cm}
    \begin{minipage}{.2\linewidth}
        \centering
        \includegraphics[width=\linewidth, scale=.15]{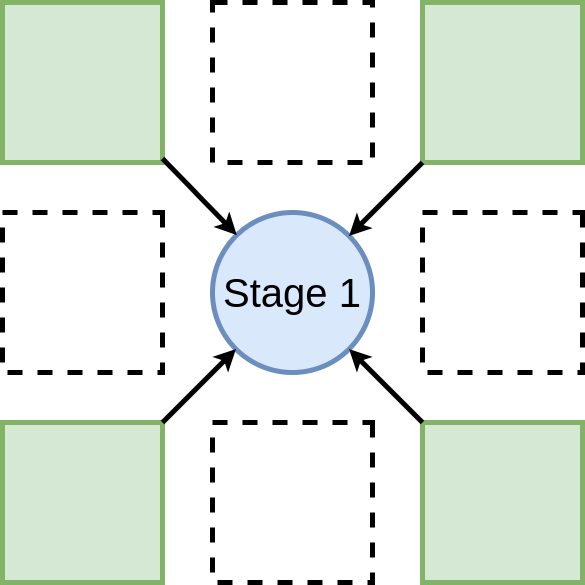}
        \subcaption{\small{4D diagonal}}
    \end{minipage}
    \hspace{0.5cm}
    \begin{minipage}{.2\linewidth}
        \centering
        \includegraphics[width=\linewidth, scale=.15]{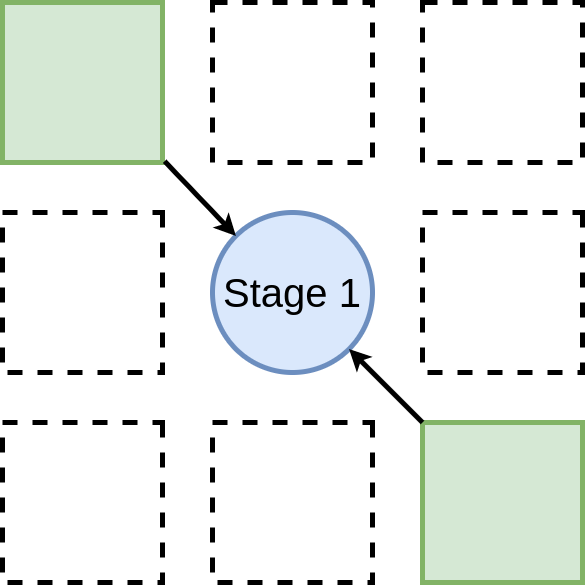}
        \subcaption{\small{2D left diag}}
    \end{minipage}
    \hspace{0.5cm}
    \begin{minipage}{.2\linewidth}
        \centering
        \includegraphics[width=\linewidth, scale=.15]{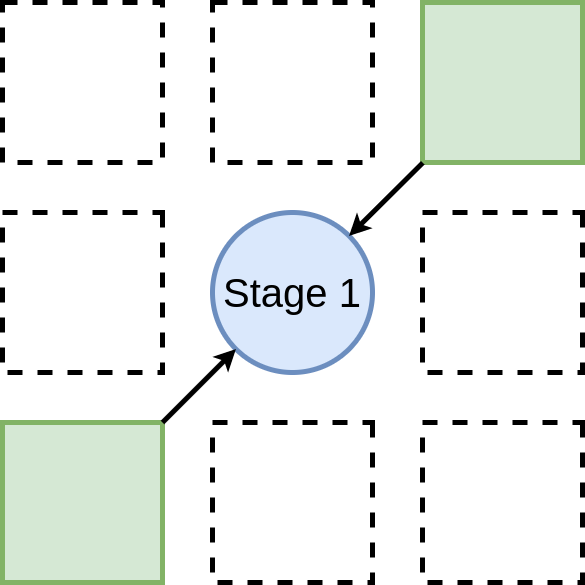}
        \subcaption{\small{2D right diag}}
    \end{minipage}
    \caption{\small{Six reconstruction strategies to interpolate 3x3 views from 4 input corner views (green), dashed square views are not used for interpolation, different colors of circles identify output views from different stages}}
	\label{fig:corners}
\end{figure}

\begin{figure}[t]
    \hspace{1.3cm}
    \begin{subfigure}{0.25\textwidth}
        \centering
        \includegraphics[width=\linewidth]{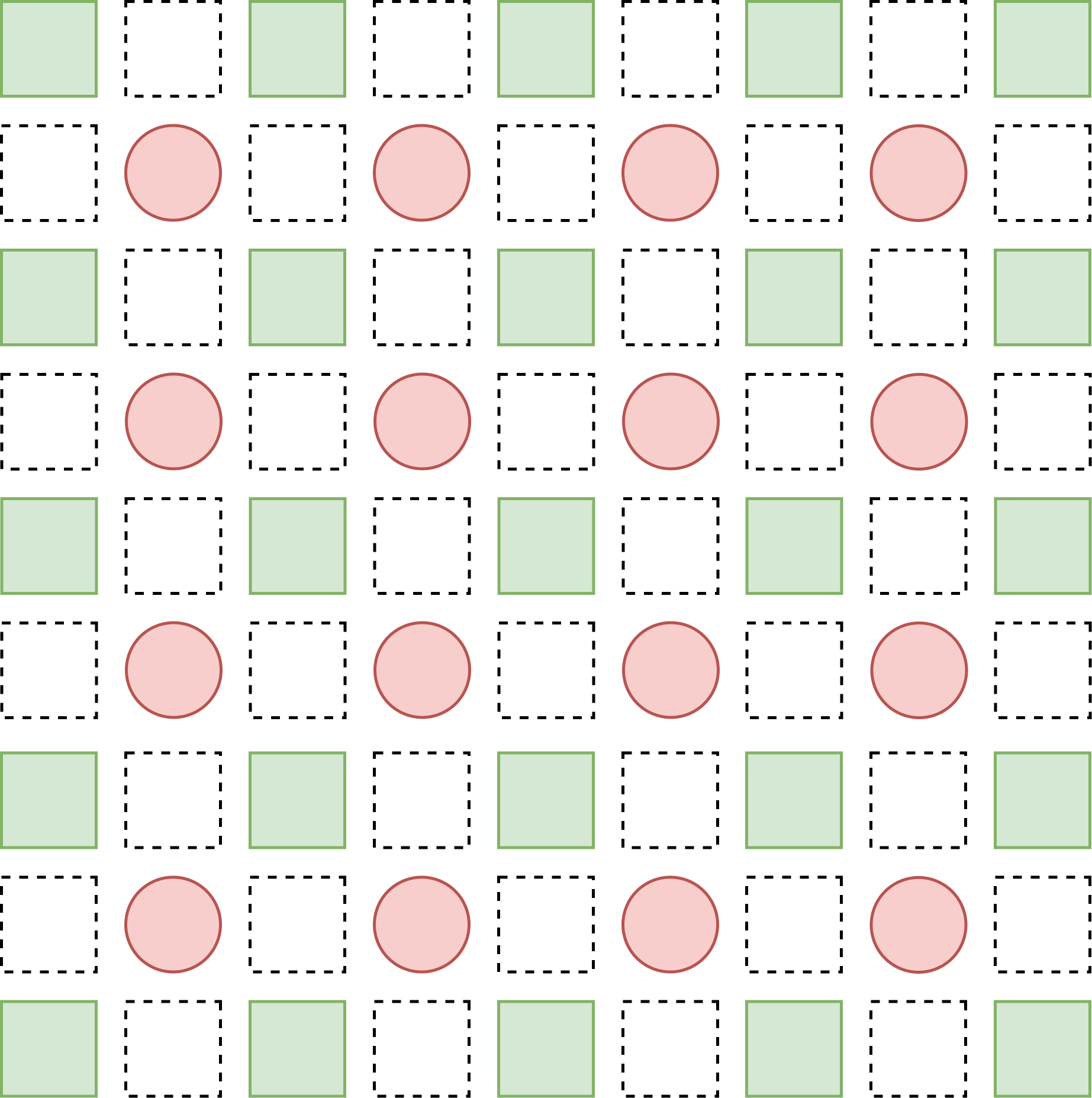} 
        \caption{\small{level 1}}
        \label{fig:sparsity_subimg1}
    \end{subfigure}
    \hspace{0.5cm}
    \begin{subfigure}{0.25\textwidth}
        \centering
        \includegraphics[width=\linewidth]{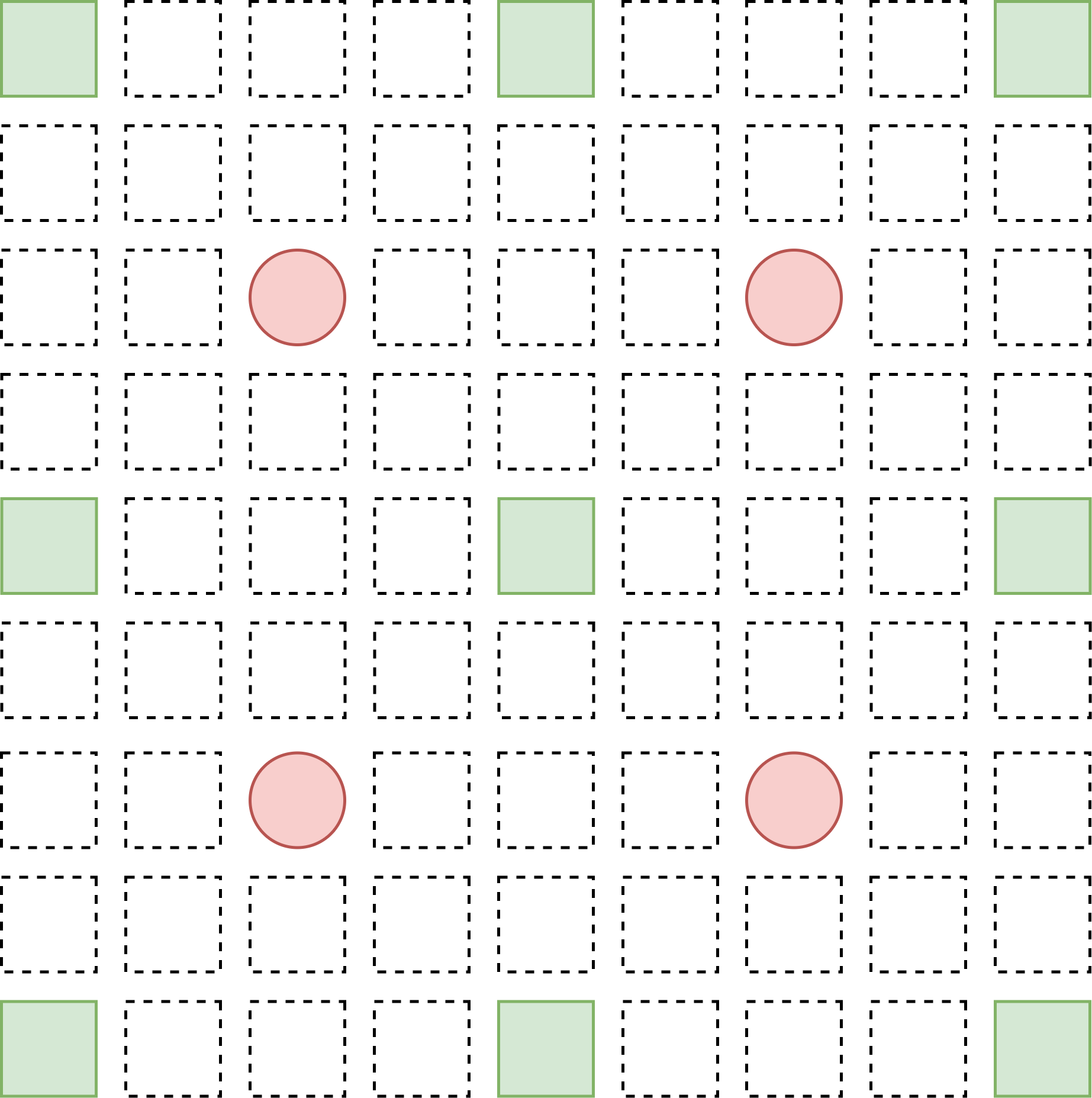}
        \caption{\small{level 2}}
        \label{fig:sparsity_subimg2}
    \end{subfigure}
    \hspace{0.5cm}
    \begin{subfigure}{0.25\textwidth}
        \centering
        \includegraphics[width=\linewidth]{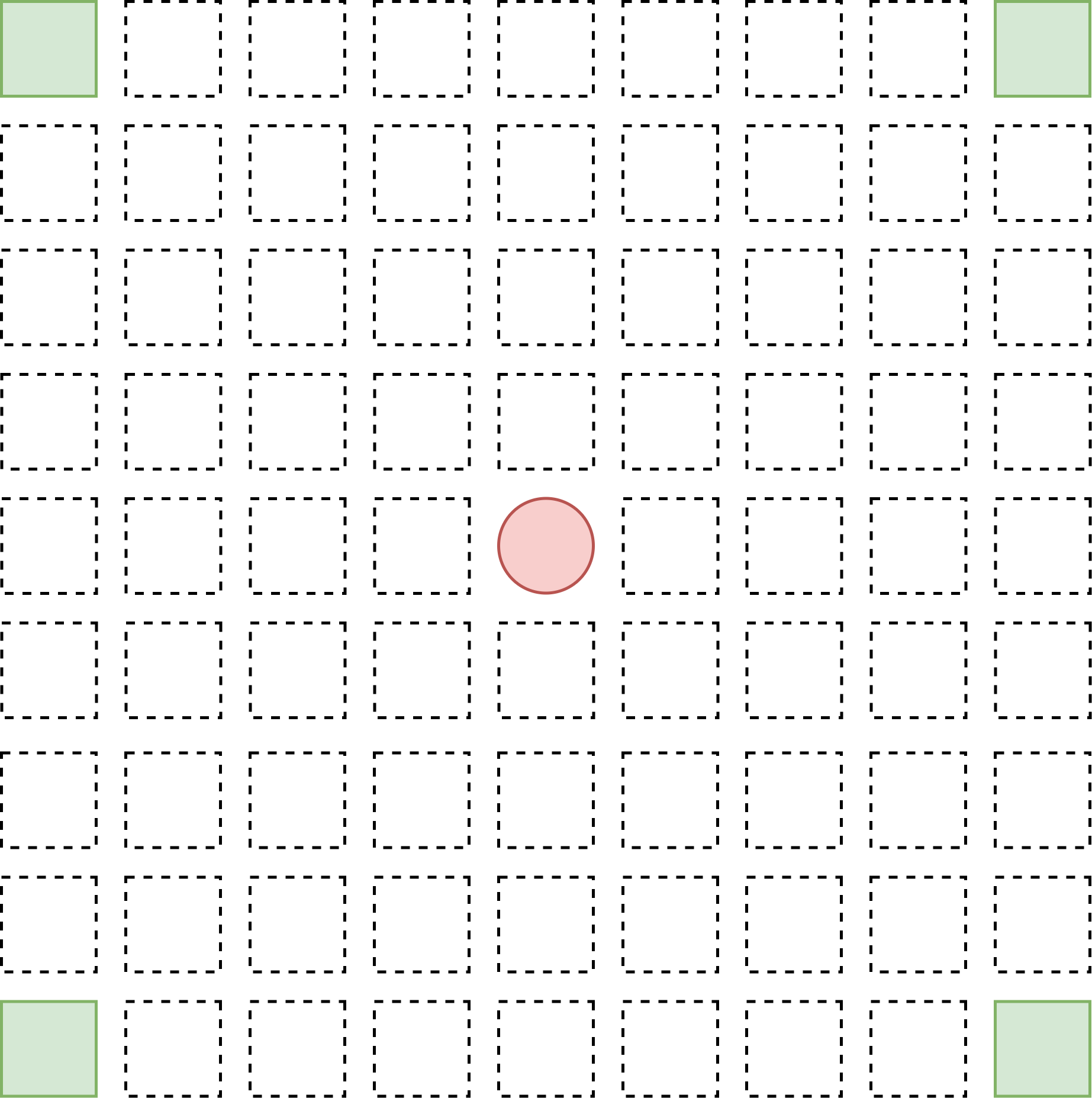}
        \caption{\small{level 3}}
        \label{fig:sparsity_subimg3}
    \end{subfigure}
     
    \caption{\small{Three levels of angular density for LF reconstruction. The $IR$s of these levels are 30.9\%, 11.1\% and 4.9\% respectively.}}
    \label{fig:sparsity}
    \vspace{-0.4cm}
\end{figure}

In this section, we investigate different strategies for light field subsampling and reconstruction. Firstly, a benchmark LF view interpolation method has to be selected from SOTA approaches to evaluate all strategies. Next, given a fixed sampling ratio, three light field subsampling strategies are studied to reconstruct full size LFs from each sampled LF (Figure~\ref{fig:fullLF}). Finally, six different reconstruction strategies are explored to generate a dense light field from inputs of varying sparsity (Figures~\ref{fig:corners} \&~\ref{fig:sparsity}).

\subsection{Benchmark method selection}
\label{subsec:SOTA}

As our goal is to investigate sampling and reconstruction strategies for LFs, we first select a benchmark method to be applied in our experiments. The benchmark has to be flexible to work in different configurations but should also provide best possible interpolation quality. We therefore evaluate SepConv~\cite{niklaus2017video}, Shearlet~\cite{vagharshakyan2017light} and LFEPI~\cite{wu2017epi} in an initial study. LFEPI is a representative learning based LF view synthesis method, and Shearlet is an efficient non-learning based reconstruction method in the Fourier domain. SepConv~\cite{niklaus2017video} was initially designed for video frame interpolation and employs a neural network based kernel estimator to interpolate views between adjacent input views. As such it is very flexible and can be also be used in various ways of LF view interpolation.

These SOTA methods are evaluated by using the same input pattern shown in Figure~\ref{fig:fullLF_subimg1} resulting from row-wise sampling, in which sampled input views are represented as green squares and reconstructed output views are represented as red circles. According to Table~\ref{tab:SOTA}, SepConv numerically outperforms all other methods significantly.
Shearlet achieves better performance than linear interpolation. LFEPI scores well on PSNR, while Shearlet performs better in terms of average SSIM. 
However, both these two methods require an EPI as input. Thus SepConv is not only the best performing approach, but also the only one that can easily be adapted to different configurations, as it only needs a pair of RGB images as input. We therefore continue to use SepConv in our further experiments.

Since SepConv was originally trained for video frame interpolation, we additionally fine-tuned the pretrained model on our light field training dataset in order to further improve the performance. Table~\ref{tab:SepConv} shows improvements we can achieve by fine-tuning and retraining the initial network. For some of our experiments detailed below, we have to retrain SepConv appropriately in order to work with more than 2 input views, e.g. left, right, top and bottom neighbors.

\subsection{Study of basic light field subsampling strategies}
\label{subsec:subsampling}

In this set of experiments, we compare basic subsampling strategies as illustrated in Figure~\ref{fig:fullLF}, with the goal to identify the most efficient basic subsampling strategy among row-wise, column-wise and checkerboard. To evaluate these strategies, the sampled LFs are reconstructed using view interpolation. After reconstruction, the quality of the syntesized views can be compared to the corresponding ground truth. All these strategies have the same ratio of sampled views to total views, which is an important measure of the sparsity and defined as the $InputRatio$ ($IR$) in equation~\eqref{eq:IR}:

\begin{equation}
IR = \frac{N_{InputViews}}{N_{InputViews}+N_{OutputViews}}
\label{eq:IR}
\end{equation}

where the numbers of input views and output views of the interpolation method are $N_{InputViews}$ and $N_{OutputViews}$, respectively. The total number of views of the completed light field can be represented as the sum of $N_{InputViews}$ and $N_{OutputViews}$.
The three basic subsampling strategies as shown in Figure~\ref{fig:fullLF} all have $IR \approx 55\%$.

We applied our fine-tuned SepConv as explained in Section~\ref{subsec:SOTA} to interpolate the necessary views for row-wise and column-wise strategies from neighbouring views. For the checkerboard pattern we had to modify the original SepConv network in order to accept 4 views as input, top, bottom, left and right. The outer views depicted as blue circles in Figure~\ref{fig:fullLF_subimg3} were synthesized by row-wise or column-wise interpolation from 2 neighbouring views.

\begin{figure}[t]
	\centering
	\includegraphics[width=0.5\linewidth]{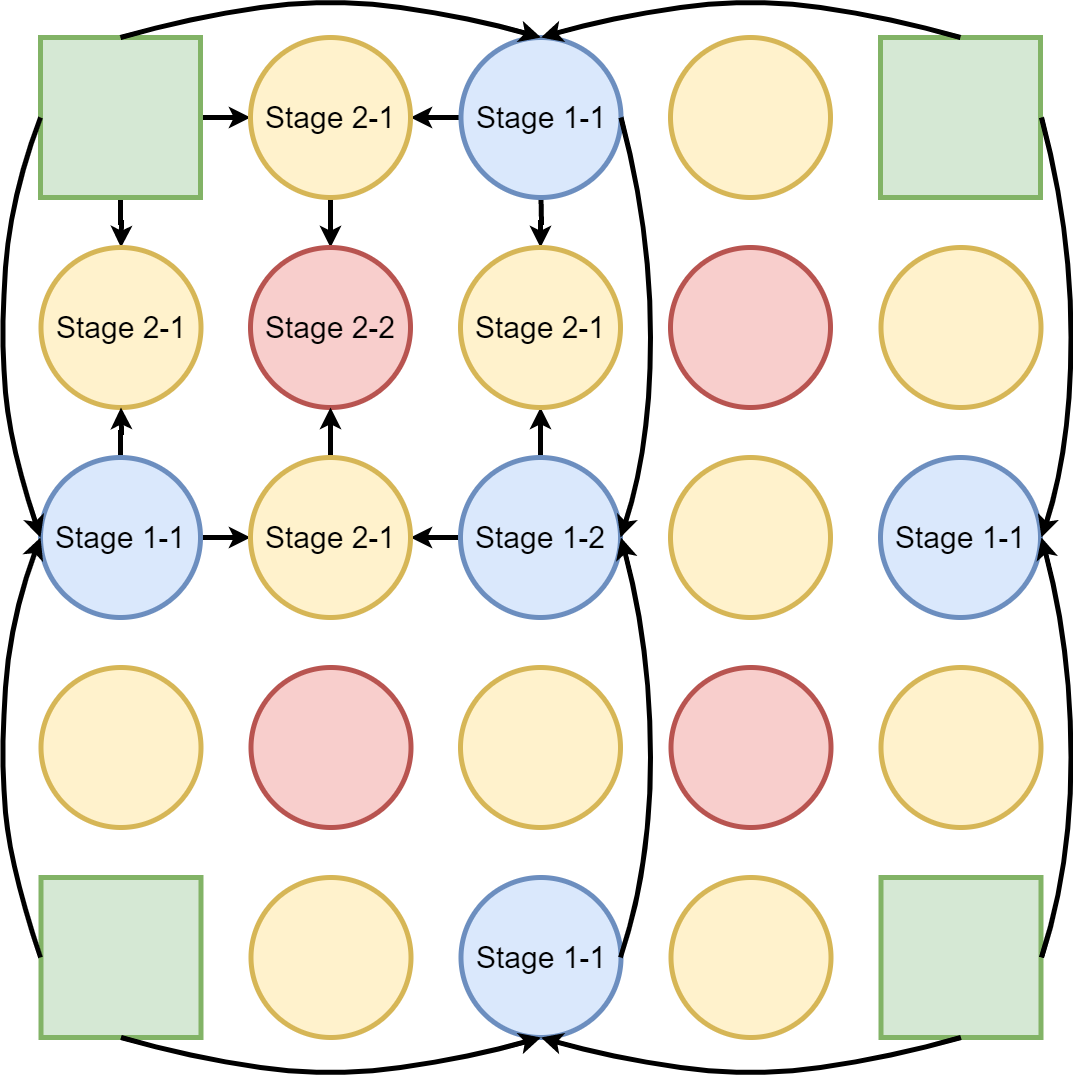}
	\caption{\small{Stages to complete a quarter of the full LF for an input angular density of level 2 with $IR = 11.1$\% (see Fig. \ref{fig:sparsity})}}
	\label{fig:complete}
	\vspace{-0.4cm}
\end{figure}

\subsection{Study of sparse light field reconstruction strategies}
\label{subsec:reconstruction}

In this set of experiments, we compare different reconstruction strategies for sparse light fields. Taking a 3x3 matrix of views as example, six progressive reconstruction strategies can be applied as presented in Figure~\ref{fig:corners}. Using four corner images as input, we can reconstruct the side images using the same row-wise and column-wise interpolation as before. Thus, the question becomes which is the best way to reconstruct the central view. Three of these strategies, including 2D horizontal-vertical (2D H-V), 2D vertical-horizontal (2D V-H) and 4D omni, are two stage cases involving generating side views as intermediate stage. The other three, including 4D diagonal, 2D left diagonal and 2D right diagonal, require only one stage to generate the central view.

To further study the influence of the angular density of the input views, we investigate three levels of density as shown in Figure~\ref{fig:sparsity}. The distance between two input views is 1 view in level 1, and there are 3 views and 7 views distance in level 2 and 3, respectively. To compare fully reconstructed light fields, we reconstruct all missing views using appropriate strategies as follows. Level 1 is completed by row-wise and column-wise interpolation of the side views, 
using the 2D H-V method unless otherwise specified. The choice of 2D H-V as the default method is justified by its better performance demonstrated in Section~\ref{sec:results}.
For level 2 and 3, we recursively fill using lower level methods to complete the full scale LF. The reconstruction of a quarter of the LF is shown in Figure~\ref{fig:complete}, which is applied iteratively to each quarter one by one. Please find more detailed completion stages in the supplementary material~\footnote{\label{web}\url{https://v-sense.scss.tcd.ie/?p=4450}}.

\section{Results}
\label{sec:results}

In this section, we summarize the results of our experiments, which were performed on an Intel Core i7-6700k 4.0GHz CPU, while neural network refining was performed on Nvidia Titan Xp GPUs. 

The performance of strategies is evaluated on both real-world and synthetic LF datasets to validate their robustness. We used 27 real-world light fields captured by Lytro Illum cameras provided by EPFL \linebreak \cite{Rerabek2016} and INRIA~\cite{INRIADataset}, and 11 light fields from the Stanford dataset taken by a camera gantry~\cite{StanfordDataset}. As for the synthetic LF dataset, all 28 light fields from the HCI benchmark \linebreak \cite{honauer2016benchmark} were used. 10 light fields in total were selected from these datasets as the test set and the rest as training set. Additionally, 160 light fields from LF intrinsic~\cite{alperovich2018light} were added for retraining of SepConv to avoid the overfitting. All views were cropped to equal 512x512 resolution to accelerate the computation.
In this study, we use the peak signal-to-noise ratio (PSNR) and the structural similarity (SSIM) on RGB images to evaluate the algorithms numerically. For each light field, unless emphasized specifically, mean numerical results are computed over all views of the full light fields.

\begin{table}[!]
\footnotesize
\begin{minipage}{\textwidth}
    \begin{minipage}[c]{0.55\textwidth}
        \centering
        \caption{Evaluation of three basic subsampling strategies from Figure~\ref{fig:fullLF}}
        \label{tab:subsampling}
        \begin{tabular}{c|c|c|c|c|c|c}
        & \multicolumn{2}{c|}{row-wise} & \multicolumn{2}{c|}{column-wise} & \multicolumn{2}{c}{checkerboard} \\
        \hline
        & \text{PSNR} & \text{SSIM} & \text{PSNR} & \text{SSIM} & \text{PSNR} & \text{SSIM} \\
        \hline
        \text{HCI} & 40.49 & 0.9956 & 40.33 & 0.9958 & 39.24 & 0.9935 \\
        \hline
        \text{Lytro} & 39.32 & 0.9932 & 39.07 & 0.9925 & 38.67 & 0.9921 \\
        \hline
        \text{Stanford} & 40.22 & 0.9936 & 39.37 & 0.9924 & 39.47 & 0.9928 \\
        \hline
        \hline
        \text{Mean} & \color{red}{\textbf{39.74}} & \color{red}{\textbf{0.9925}} & 39.20 & 0.9915 & 39.33 & 0.9918
    \end{tabular}
    \end{minipage}
    \hspace{0.7cm}
    \begin{minipage}[c]{0.4\textwidth}
        \centering
        \caption{Evaluation of six reconstruction strategies from Figure~\ref{fig:corners}}
        \label{tab:six_corners}
        \begin{tabular}{c|c|c}
        & PSNR & SSIM \\
        \hline
        \text{2D H-V} & \color{red}{\textbf{37.42}} & \color{red}{\textbf{0.9884}} \\
        \hline
        \text{2D V-H} & 37.21 & 0.9881 \\
        \hline
        \text{2D left diagonal} & 35.77 & 0.9846 \\
        \hline
        \text{2D right diagonal} & 35.86 & 0.9842 \\
        \hline
        \text{4D omni} & 36.51 & 0.9863 \\
        \hline
        \text{4D diagonal} & 36.86 & 0.9838
    \end{tabular}
    \end{minipage}
\end{minipage}
\end{table}

The results of the evaluation of basic subsampling strategies are shown in Table~\ref{tab:subsampling}. 
Row-wise interpolation achieves the best scores on most light fields over all datasets. Please view our website for comprehensive results of each light field~\textsuperscript{\ref{web}}.
The difference to column-wise is most prominent for the Stanford data which was captured by a gantry, resulting in unequal vertical displacement.
For the checkerboard pattern, the retraining of SepConv to accept 4 views may be the reason of its worse performance, as this way it could not benefit from the pretrained model of SepConv. 

\begin{figure}[!t]
    \hspace{1.5cm}
    \begin{minipage}[c]{.2\linewidth}
        \centering
        \subcaption{\small{full size GT}}
        \includegraphics[width=\linewidth, scale=.13]{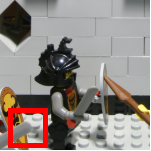}
    \end{minipage}
    \begin{minipage}[c]{.2\linewidth}
        \centering
        \subcaption{\small{2D H-V}}
        \includegraphics[width=\linewidth, scale=.15]{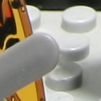}
    \end{minipage}
    \begin{minipage}[c]{.2\linewidth}
        \centering
        \subcaption{\small{2D V-H}}
        \includegraphics[width=\linewidth, scale=.15]{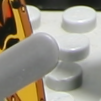}
    \end{minipage}
    \begin{minipage}[c]{.2\linewidth}
        \centering
        \subcaption{\small{4D omni}}
        \includegraphics[width=\linewidth, scale=.15]{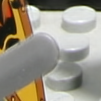}
    \end{minipage}\par\medskip
    \hspace{1.5cm}
    \begin{minipage}[c]{.2\linewidth}
        \centering
        \includegraphics[width=\linewidth, scale=.15]{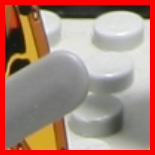}
        \subcaption{\small{GT}}
    \end{minipage}
    \begin{minipage}[c]{.2\linewidth}
        \centering
        \includegraphics[width=\linewidth, scale=.15]{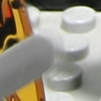}
        \subcaption{\small{4D diagonal}}
    \end{minipage}
    \begin{minipage}[c]{.2\linewidth}
        \centering
        \includegraphics[width=\linewidth, scale=.15]{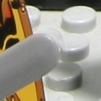}
        \subcaption{\small{left diag}}
    \end{minipage}
    \begin{minipage}[c]{.2\linewidth}
        \centering
        \includegraphics[width=\linewidth, scale=.15]{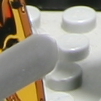}
        \subcaption{\small{right diag}}
    \end{minipage}
    \caption{\small{Visual results of six reconstruction strategies from Figure~\ref{fig:corners}}}
	\label{fig:visual_results}
	\vspace{-0.5cm}
\end{figure}

The results of sparse LF reconstruction strategies are shown in Table~\ref{tab:six_corners}. Again, 4D strategies use a retrained model while the others use a fine-tuned model.
The central view is used to evaluate these results as it is the only one that gets always reconstructed using any of the six strategies when filling a 3x3 block of views.
2D H-V performs best, as it accumulates less error than other strategies with 2 stages (row-wise first gives best reference for stage 2), while it has smaller distance between input views compared to direct diagonal strategies.
Visual results of reconstruction strategies are shown in Figure~\ref{fig:visual_results}. Occlusion artifacts around the tip of sword can be observed in diagonal strategies.

Finally, the effect of different levels of angular density is studied in Table~\ref{tab:patterns}. Since 2D H-V is the optimal strategy according to the previous conclusion, it is ultilized to complete the full scale LFs recursively as explained before. All synthesized views are averaged in this evaluation (different from only central views in Table~\ref{tab:six_corners}). From Table~\ref{tab:patterns} we get the expected decrease of interpolation quality with sparsity.

These insights about sparsity vs. quality and the best subsampling and reconstruction strategies can be beneficial for the design of LF coding approaches (maximum quality that can be achieved when omitting views) or camera systems (maximum camera distance for a desired quality and density).


\begin{table}[t]
\footnotesize
    \begin{center}
    \caption{Comparison between level 1 \& 2 \& 3 from Figure~\ref{fig:sparsity} using 2D H-V}
    \label{tab:patterns}
    \begin{tabular}{c|c|c|c}
        2D H-V & Level 1 & Level 2 & Level 3 \\
        \hline
        \text{PSNR(dB)} & 37.34 & 35.12 & 31.79 \\
        \hline
        \text{SSIM} & 0.9886 & 0.9825 & 0.9613
    \end{tabular}
    \end{center}
    \vspace{-0.4cm}
\end{table}





\section{Conclusion}
\label{sec:conclusion}

In this paper, we presented a comprehensive study comparing different strategies for efficient light field subsampling and reconstruction.
For this purpose we selected an existing view synthesis method among the best performing state-of-the-art techniques.
Using this benchmark method, we first evaluate the best subsampling approach among a row-wise, a column-wise, and a checkerboard pattern with a fixed interpolation ratio, which concludes that the row-wise approach offers the best performances.
Second, we investigate corner-based central view generation and compare the performance of six possible reconstruction strategies.
In addition, we evaluate a multi-stage approach to reconstruct dense light fields from subsampled input with different levels of angular density.
We found that using a row-wise followed by a col-wise reconstruction yields best performance.
We hope these findings will help inspire researches related to light field subsampling and reconstruction, such as compression and camera array design. 
Further, a more explicit analysis regarding the relation of the disparity range to the strategy selection could be performed.






\bibliographystyle{apalike}
\bibliography{imvip}


\end{document}